\documentclass{article}

\usepackage[nonatbib, preprint]{neurips_2020}

\usepackage[utf8]{inputenc} 
\usepackage[T1]{fontenc}    
\usepackage{hyperref}       
\usepackage{url}            
\usepackage{booktabs}       
\usepackage{amsfonts}       
\usepackage{nicefrac}       
\usepackage{microtype}      
\usepackage{graphicx}       
\usepackage{color, colortbl} 
\usepackage{multirow} 
\usepackage{longtable}
\usepackage[most]{tcolorbox} 
\usepackage{lipsum} 
\usepackage[utf8]{inputenc}
\usepackage{tcolorbox}
\usepackage{amsmath}
\usepackage{tcolorbox}

\usepackage{booktabs} 
\usepackage{tabularx} 
\usepackage{geometry} 

\usepackage[utf8]{inputenc}
\usepackage{enumitem}

\definecolor{headercolor}{gray}{0.85}
\definecolor{maxvalue}{rgb}{0.92,1,0.92} 

\tcbuselibrary{skins}

\title{Introducing L2M3, A Multilingual Medical Large Language Model to Advance Health Equity in Low-Resource Regions
}

\author{
  Agasthya Gangavarapu\\
  \texttt{august@uheal.ai} 
}

\begin{document}

\maketitle

\begin{abstract}

Addressing the imminent shortfall of 10 million health workers by 2030, predominantly in Low- and Middle-Income Countries (LMICs), this paper introduces an innovative approach that harnesses the power of Large Language Models (LLMs) integrated with machine translation models. This solution is engineered to meet the unique needs of Community Health Workers (CHWs), overcoming language barriers, cultural sensitivities, and the limited availability of medical dialog datasets. I have crafted a model that not only boasts superior translation capabilities but also undergoes rigorous fine-tuning on open-source datasets to ensure medical accuracy and is equipped with comprehensive safety features to counteract the risks of misinformation.

Featuring a modular design, this approach is specifically structured for swift adaptation across various linguistic and cultural contexts, utilizing open-source components to significantly reduce healthcare operational costs. This strategic innovation markedly improves the accessibility and quality of healthcare services by providing CHWs with contextually appropriate medical knowledge and diagnostic tools. This paper highlights the transformative impact of this context-aware LLM, underscoring its crucial role in addressing the global healthcare workforce deficit and propelling forward healthcare outcomes in LMICs.

\end{abstract}

\section{Introduction}

The COVID-19 pandemic has starkly highlighted the vulnerabilities of healthcare systems in low-resource countries, intensifying the strain on these systems. The inadequacy of traditional healthcare delivery models, especially in rural areas plagued by infrastructural deficits, workforce shortages \cite{AgyemanManu2023Prioritising}, and geographic isolation, has necessitated a pivot towards more flexible and accessible forms of healthcare provision. Community Healthcare Workers (CHWs), despite their pivotal role in bridging the healthcare access gap, often face challenges due to a lack of formal training, recognition, and support, which impairs the quality and safety of the care they provide.
The potential of Large Language Models (LLMs) to augment the capabilities of CHWs, offering crucial support in diagnostics, primary care delivery, and clinical decision-making, represents a significant advancement. Despite the potential of LLMs in healthcare, their deployment for CHWs in Low- and Middle-Income Countries (LMICs) faces significant hurdles \cite{Ahmed2022}. Efforts to tailor LLMs for these settings are scarce, and the challenges are manifold. They range from modifying LLMs to deliver precise medical guidance, overcoming linguistic hurdles for worldwide utility, to adapting the models to account for cultural nuances and the particular health needs of different regions. Furthermore, there are concerns regarding an excessive dependency on LLMs, the reliability of the medical advice dispensed, and the economic feasibility of implementing such advanced technologies in environments where resources are constrained, all of which hinder their adoption within healthcare frameworks  \cite{medicalailab}.

This paper proposes a comprehensive development strategy to refine LLMs for healthcare use in low-resource environments. By integrating sophisticated machine translation technologies, implementing advanced retrieval techniques, and establishing a mechanism for continuous feedback from CHWs and healthcare professionals, the proposed approach aims to address the linguistic, cultural, and contextual challenges associated with LLM deployment. Enhancing the accuracy, reliability, and cultural relevance of LLMs through this integrated methodology will empower CHWs, facilitating improved healthcare delivery and access in underserved areas.

\section{CHWs and LLMs in Rural Areas}
In developing nations across Africa and Asia, the acute shortage of healthcare professionals, including doctors, nurses, and specialists, is a critical issue. CHWs are essential in these regions, with India alone employing over a million Accredited Social Health Activists (ASHA \cite{ashaindia}) and Auxiliary Nurse Midwives (ANM). Despite CHWs' effectiveness, challenges such as inadequate training, poor compensation, and burnout hinder their service delivery. Additionally, there's a pronounced global shortage of healthcare workers; for example, India's healthcare worker density is significantly below WHO's recommended levels, a situation echoed in many African countries. This shortage poses a major obstacle to achieving healthcare equity, emphasizing the need for significant investment in healthcare workforce development.
This integration, particularly vital in rural or remote locales, mandates a specialized form of training and support for CHWs. The principal challenges faced by CHWs include:

\begin{itemize}
    \item The geographical and financial inaccessibility of formal training institutions, which are predominantly situated in distant urban centers, poses a significant barrier to the acquisition of formal education and training for CHWs.
    \item The requirement for extended training periods often translates into a potential loss of income for CHWs, exacerbating the financial strain and deterring participation in necessary educational programs.

    \item The conventional, physician-centric model of healthcare delivery is relatively impractical in these settings due to the requisite investment in extensive building and infrastructure, which contrasts with the inherently mobile nature of CHWs who traverse diverse locations to reach their patients.

\end{itemize}
LLMs such as ChatGPT introduces a significant advancement in addressing the educational and operational exigencies confronted by CHWs in LMICs. The capacity of LLMs to emulate human-like textual interactions presents an innovative avenue for augmenting both the educational framework and the medical assistance provided to CHWs, thereby facilitating a more effective healthcare delivery system in underserved areas. The potential applications and benefits of LLMs in this context are multifaceted:
\subsection{Medical Assistant:}
\begin{itemize}
    \item \textbf{Decision Support:} LLMs can serve as an invaluable decision support mechanism for CHWs, offering instant guidance on diagnostic techniques, treatment plans, and patient care strategies, particularly in regions where medical professionals are scarce.
    \item \textbf{Information Retrieval:} The ability of CHWs to access up-to-date medical information, guidelines, and evidence-based practices through LLMs ensures that their interventions are aligned with contemporary medical standards and medicine adverse reactions.
    \item \textbf{Health Data Reporting:} LLMs can aid CHWs in efficiently gathering and communicating health information, thereby enhancing the effectiveness of community health initiatives through improved accuracy and timeliness of health data reporting from rural and remote areas to central health systems.

\end{itemize}

\subsection{Educational Tool:}
\begin{itemize}
    \item \textbf{Personalized Learning Experiences:}LLMs can tailor educational content to meet the specific needs of CHWs and address the unique health challenges prevalent within their communities. This personalized approach to learning is pivotal in surmounting the educational barriers posed by the physical and financial inaccessibility of traditional academic institutions.
    \item \textbf{Adaptability to Language and Literacy Levels:}The linguistic diversity inherent in LMICs necessitates the provision of training material in multiple languages. LLMs can dynamically adjust the complexity of educational content to match the literacy levels of CHWs, thus democratizing access to learning.
    \item \textbf{Engagement through Interactive Learning:}Employing interactive formats such as Q\&A sessions, simulated scenarios, and problem-based learning, LLMs can significantly enhance the engagement and practical application skills of CHWs, fostering a deeper understanding and retention of medical knowledge.

\end{itemize}

The potential of LLMs to enhance the efficacy of CHWs in LMICs is substantial. However, a critical examination reveals that the majority of LLMs, including advanced platforms like ChatGPT and MedPaLM 2, are predominantly designed with the infrastructural and cultural contexts of developed nations and urban environments in mind \cite{Biderman2021DatasheetPile}. This focus has contributed to previous AI-driven healthcare initiatives falling short of significantly reducing health inequities in LMICs \cite{OatmealHealth2023}.

\section{Uheal L2M3 Model System}

\subsection{Design Goals}
The Uheal L2M3 Model System is designed to empower CHWs effectively and safely, and aimed at tackling key health issues outlined in the WHO's Sustainable Development Goals for LMICs. Key design considerations are:
\begin{itemize}

\item \textbf{Simplified Communication:} Ensures CHWs receive clear, accessible diagnostic and educational feedback, avoiding complex medical terms and abbreviations for better understanding.
\item \textbf{Public Health Priorities:} Focuses on reducing Disability-Adjusted Life Years (DALYs) by aligning with WHO-highlighted health conditions in SDGs for LMICs.
\item \textbf{Content Safety and Accuracy:}Integrates guardrails to review inputs and outputs for toxicity, misinformation, or misguidance, ensuring reliable and safe information exchange.
\item \textbf{Modularity and Scalability:} Features a flexible architecture that adapts to various geographical and cultural settings with minimal adjustments, enhancing widespread applicability.
\item \textbf{Cultural Sensitivity and Localization:} Prioritizes cultural nuances and localization, customizing content to fit local languages, customs, and health beliefs. This approach fosters culturally relevant and effective interventions, improving health outcomes.

\end{itemize}

\subsection{Development Process of the Uheal L2M3 Model}
To construct the Uheal L2M3 model system, a systematic methodology was employed, guided by predefined design objectives. The process commenced with the compilation of a specialized training corpus, incorporating medical dialogue datasets alongside guidelines issued by regional health authorities. This initial step ensured the model's relevance and applicability to healthcare communications.

Subsequently, an evaluation of various domain-adapted LLMs was conducted. This assessment focused on several key attributes: the breadth and specificity of the models' pre-trained datasets, their multilingual capabilities, and their pertinence to CHWs. Based on a set of established criteria, two models were selected for further development. These models were then fine-tuned using the assembled dialog corpus and guidelines, enhancing their performance and specificity to the target domain.

Additionally, the integration of machine translation models was explored to extend the system's applicability across different linguistic regions, thereby addressing the challenge of scalability and regional language context. This step involved evaluating various translation models for their accuracy and effectiveness in medical dialogues.

The fine-tuning of the model systems was followed by a comprehensive evaluation phase. This phase employed multiple methodologies to rigorously assess the models' performance, with a particular focus on ensuring their safe and secure deployment in healthcare settings. These methodologies encompassed both quantitative and qualitative analyses, aiming to validate the models' effectiveness, reliability, and adherence to healthcare communication standards.

\subsection{Medical Training and Evaluation Data}
The Uheal L2M3 System leverages a domain-adaptive fine-tuning approach, incorporating a comprehensive medical corpus that amalgamates 930 million tokens sourced from six distinct datasets. This corpus includes clinical guidelines and guidelines provided by regional health authorities. Additionally, it encompasses medical dialogue datasets, which consist of conversations between healthcare providers and patients, as well as adverse events data sourced from the FDA Adverse Event Reporting System (FAERS). 
\begin{figure}[htp]
    \centering
    \includegraphics[scale=0.32]{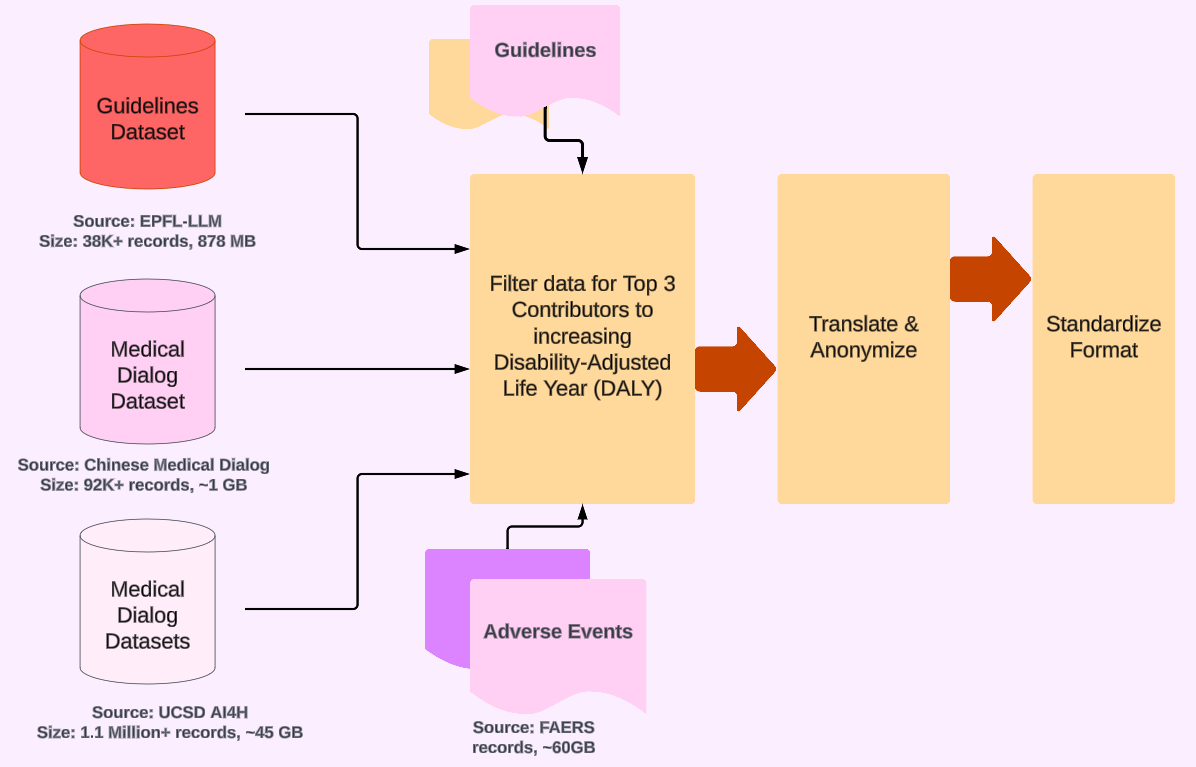}
    \caption{Data Acquisition and Standardization }
    \label{fig:graphic1}
\end{figure}

The selection and filtration of these datasets were strategically guided by a focus on the three primary contributors to Disability Adjusted Life Years (DALYs) \cite{WHO2023IMRDetails}: Ischemic Heart Disease (IHD), Lower Respiratory Infections (LRIs), and Neonatal Care. DALYs, a comprehensive metric for assessing the burden of disease, serve as a crucial indicator for evaluating healthcare delivery, especially in LMICs. This deliberate emphasis on DALYs allowed for targeted training on areas with significant health impacts, ensuring that the approach not only aligns with global health priorities but also contributes effectively to addressing the most pressing healthcare challenges in LMICs.

\begin{table}[h!]
  \centering
  \begin{tabular}{lcccc}
    \toprule
    \textbf{Dataset} & \multicolumn{2}{c}{\textbf{Number of samples}} & \multicolumn{2}{c}{\textbf{Number of related samples}} \\
    \cmidrule(r){2-3} \cmidrule(l){4-5}
     & Train & Validation & Train & Validation \\
    \midrule
    Medical Dialogues \cite{UCSDAI4H2023} & 1.1M & 15K (1.3\%) & 420K & 7K (1\%) \\
    Clinical Guidelines \cite{EPFLLLMGuidelines} & 41K & 2284 (5\%) & 9K & 1180 (12\%) \\
    Medical Dialogues \cite{Toyhom2023} & 92K & 4K (4\%) & 19K & 640 (6\%) \\
    Regional Guidelines & 2.4K & 0 (0\%) & 1.4K & 600 (0\%) \\
    Medical Adverse Events \cite{FDAFAERS2023}&  61K & 3K (5\%) & 5K & 512 (10\%) \\
    \bottomrule
  \end{tabular}
   \vspace{10pt} 
  \caption{The size of both training and validation sets for fine-tuning. Both the medical dialog datasets are in Chinese and translated after filtering}
  \label{tab:your_label}
\end{table}

\subsubsection{Translation}
The datasets I've gathered and curated predominantly consist of English-language materials, except for two dialogue datasets that are in Chinese. To translate these dialog datasets, I utilized the Azure AI Translator to translate the medical dialogue datasets into multiple languages: English, Telugu, Hindi, Arabic, and Swahili. However, I observed that some of the translated terms were rendered literally, posing a potential challenge for CHWs due to the lack of colloquialisms. These literal translations could lead to confusion, as they might not align with the everyday language used by the target audience. To address this issue and ensure the translations are both culturally appropriate and easily understandable, I implemented post-editing techniques. This involved carefully replacing the directly translated terms with phrases and words that are more frequently used and culturally sensitive, thereby enhancing the accessibility and relevance of the translated content for CHWs.

 \subsubsection{Anonymization}
Although the datasets utilized in this study are sourced from publicly accessible repositories and are presumed to be devoid of any Personally Identifiable Information (PII), it was imperative to verify the absence of any confidential data within them. To address this concern, I implemented a three-shot prompting technique through the GPT-4 API, designed to identify PII such as Names, email addresses, etc., and anonymize those elements in the datasets. I found no PII or sensitive personal information in any of the datasets.  

\subsubsection{Data Formats}

 Curated datasets are formatted using OpenAI ChatML \cite{openai2023chatml} format for instruction fine-tuning.  ChatML documents consist of a series of messages, starting with a special token
\textit{<|im\_start|>}, followed by the role of messenger (i.e., the “user” or the “assistant”), a new line,
and then the message itself. The message is then suffixed with a second special token: \textit{<|im\_end|>}.  

\begin{table}[h!]
  \centering
  \begin{tabular}{>{\bfseries}l p{11cm}}
    \toprule
    Dataset & \textbf{Format} \\
    \midrule
    Medical Dialog Dataset & \textbf{< |im\_start| > CHW}
    I have a case where a 2-week-old baby presents with yellowing of the skin and eyes. The parent is concerned. Can you provide some guidance on what might be causing this? 
    \textbf{< |im\_end| >}

    \textbf{< |im\_start| > Assistant} 
    The symptoms described are indicative of jaundice, which is common in newborns. Jaundice in infants .. \textbf{< |im\_end| >} 
    
    \textbf{< |im\_start| > CHW}
    What initial advice should I give the parent?
    \textbf{< |im\_end| >}

    \textbf{< |im\_start| > Assistant} 
    Advise the parent to ensure the baby is feeding well, whether breastfed or formula-fed, as this can help lower bilirubin levels. They ......level. \textbf{< |im\_end| >} \\
  \midrule
  Guidelines & \textbf{< |im\_start| > CHW}
    What are the guidelines for malnutrition in children?
    \textbf{< |im\_end| >}

    \textbf{< |im\_start| > Assistant} 
    WHO guidelines for assessment of malnutrition.. \textbf{< |im\_end| >} \\

    \bottomrule
  \end{tabular}
  \vspace{10pt} 
  \caption{Datasets prepared for instruction fine-tuning and the format}
  \label{tab:your_label}
\end{table}

\subsection{Foundational Model Evaluation}

The development and training of foundational AI models, such as GPT-4 and Llama 2, leveraging an expansive corpus of medical data from sources like PubMed, have shown promising results. These models have exhibited proficiency in responding to general prompts across a variety of non-English languages, underscoring their potential for broad application. However, the capacity of these models to deliver reliable medical advice in low-resource languages, including Telugu and Bengali, requires further investigation. My research focuses on using pre-trained LLMs in medical settings, aiming to improve healthcare delivery by overcoming language barriers, particularly in regions with limited resources. This effort is guided by two main strategies designed to maximize the potential of these sophisticated AI tools.

Firstly, I investigated the direct application of these AI models in interpreting medical dialogues across various languages, assessing their ability to provide reliable advice without any language-specific or domain-specific fine-tuning. Secondly, recognizing the limitations of current models in comprehensively understanding and responding to complex medical inquiries in low-resource languages, I explored the option of integration of effective machine translation tools to adapt prompts into low-resource languages, followed by leveraging domain-specific trained models in English. This strategy not only circumvents the challenges posed by language barriers but also enriches the models' responses with the nuanced understanding necessary for medical advisement.

\subsubsection{Multilingual Evaluation of LLMs}

The first approach involved evaluating model performance across curated medical dialog datasets in four languages—Telugu, Hindi, Swahili, and Arabic—by measuring the semantic similarity between the generated answers and human-edited counterparts. The long-form text responses, generated by LLMs, present unique challenges in evaluating using more common metrics like the Recall-Oriented Understudy for Gisting Evaluation (ROUGE) metric. This is primarily because the ROUGE approach does not effectively capture context-sensitive information, a crucial component in evaluating semantic congruence. Hence evaluated the responses using multilingual BERT (mBERT). This model leverages contextualized embeddings to better understand the nuanced meanings embedded within the text, offering a more refined analysis of semantic alignment between model-generated responses and reference answers.

\begin{figure}[htp]
    \centering
    \includegraphics[scale=0.20]{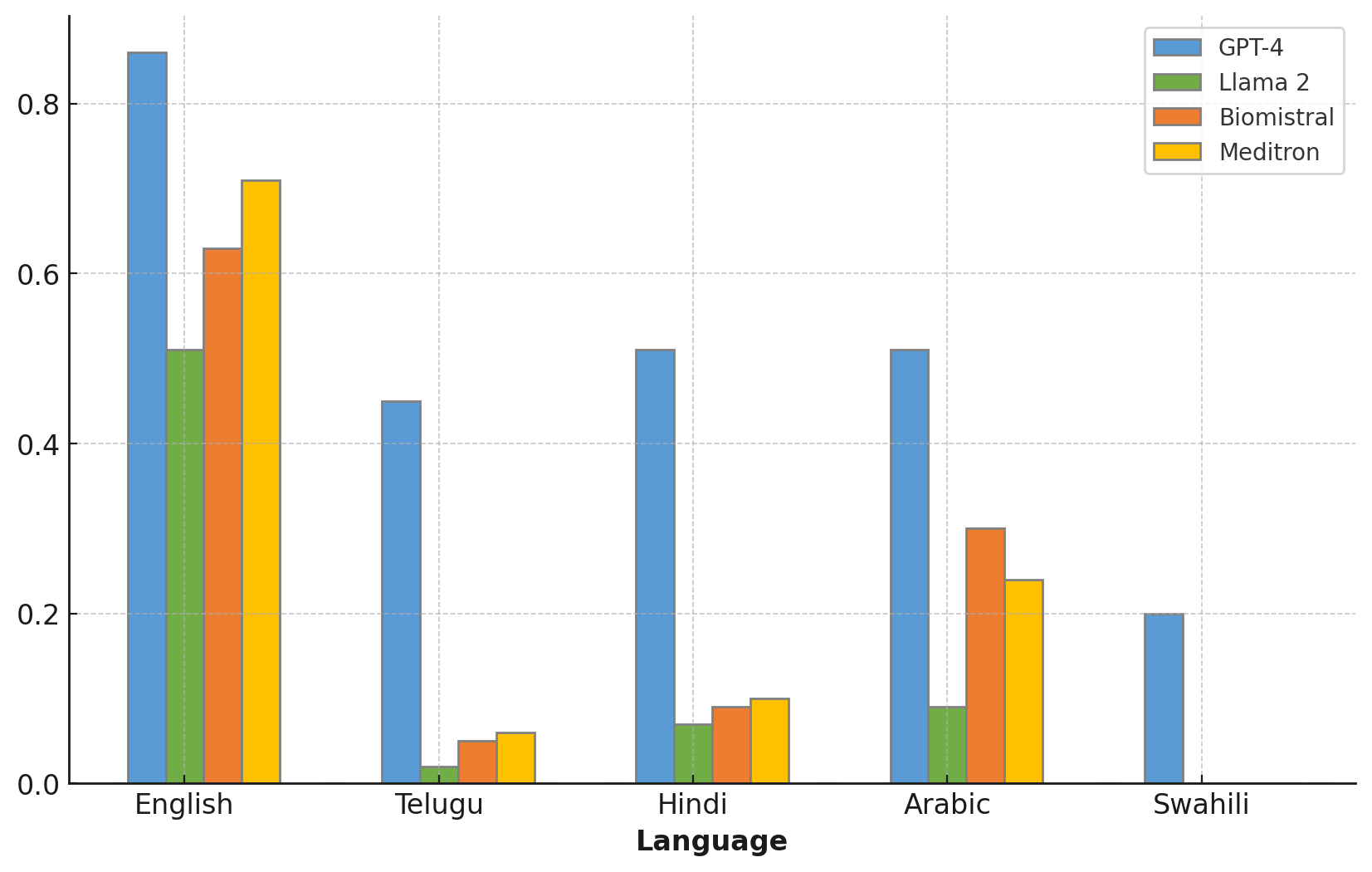}
    \caption{Comparative Performance of GPT-4, Llama 2,  Biomistral, and Meditron }
    \label{fig:graphic1}
\end{figure}

The semantic similarity chart (Figure 2) comparing GPT-4 \cite{gpt4systemcard}, Llama 2 \cite{touvron2023llama}, Biomistral \cite{labrak2024biomistral}, and Meditron \cite{chen2023meditron70b} across English, Telugu, Hindi, Arabic, and Swahili reveals key insights into these models' language understanding. GPT-4 outperforms in English, but shows a significant decrease in performance for non-English languages. Llama 2, Biomistral, and Meditron struggle significantly with non-English languages. This analysis shows both the progress and limitations in current language technology, especially in achieving comprehensive multilingual understanding in domains like healthcare. 

\subsubsection{Integrated Approach with a Machine Translation Model}

In the revised approach, the emphasis shifts towards evaluating an integrated framework that harnesses the strengths of two distinct models, rather than relying solely on a single LLM. This dual-model system is strategically designed to separate and specialize in two critical functions: machine translation and answer generation. By adopting this bifurcated approach, the investigation aims to explore the synergistic potential that arises from combining specialized models, each excelling in their respective domains. This nuanced evaluation seeks to uncover the extent to which a collaborative model architecture can surpass the capabilities of standalone LLMs in delivering precise translations and contextually accurate answers. Having conducted a thorough evaluation of the English language capabilities across various LLMs, the focus now shifts towards an in-depth assessment of translation accuracy using open-source models, including but not limited to IndicTrans2 \cite{gala2023indictrans}.

\begin{table}[h!]
\centering

\begin{tabular}{>{\bfseries}lcccc}
\toprule

& GPT-3.5 & Helsinki-NLP \cite{HelsinkiNLP2023}& Seamless M4T & IndicTrans2 \\
\midrule
Telugu \textrightarrow{} English & 16.7 & 19.9 & \cellcolor{maxvalue}75.6 & 56.7 \\
Hindi \textrightarrow{} English & 16.9 & 22.1 & \cellcolor{maxvalue}73.4 & 64.7 \\
Swahili \textrightarrow{} English & 15.6 & 17.8 & \cellcolor{maxvalue}45.8 & - \\
Arabic \textrightarrow{} English & 19.3 & 24.5 & \cellcolor{maxvalue}68.5 & - \\
\midrule
English \textrightarrow{} Telugu & 11.7 & 16.8 & \cellcolor{maxvalue}59.1 & 61.8 \\
English \textrightarrow{} Hindi & 19.4 & 20.1 & \cellcolor{maxvalue}62.1 & 66.3 \\
English \textrightarrow{} Swahili & 16.6 & 19.8 & \cellcolor{maxvalue}32.4 & - \\
English \textrightarrow{} Arabic & 21.4 & 14.6 & \cellcolor{maxvalue}54.3 & - \\
\bottomrule
\end{tabular}
\vspace{10pt} 
\caption{Machine Translation Model Performance on Curated Datasets}
\end{table}

Table 2 presents a comparative analysis of the performance metrics (using BLEU) across various translation models engaged in distinct translation tasks. Notably, Meta's Seamless M4T v2 Large \cite{TheVerge2023MetaAI} exhibits superior performance across the entire spectrum of translation activities under consideration. Moreover, the expansive linguistic capabilities of Seamless M4T V2 Large, encompassing machine translation support for over 200 languages, positions it as a highly scalable foundation for any integrated system.

In the examination of the integrated system combining Meditron with Meta's Seamless M4T v2 Large, the derived accuracy metric for Telugu translations stands at approximately \textit{0.48 (0.71 * 0.675}. This preliminary outcome suggests a foundational efficacy, yet it also highlights the substantial potential for enhancement through fine-tuning the models with domain-specific datasets and the optimization of embeddings. Furthermore, the complexity of an integrated model necessitates a thorough investigation into error propagation dynamics and potential sources of loss throughout the system. 

\begin{alignat}{2}
& A_{\text{overall}} &&< A_{\text{trans}} \times A_{\text{LM}} \quad \text{where} \\
& A_{\text{overall}} &&: \textit{Overall performance or effectiveness} \nonumber \\
& A_{\text{trans}} &&: \textit{Translation component's performance} \nonumber \\
& A_{\text{LM}} &&: \textit{Language Model's performance} \nonumber
\end{alignat}

Such an analysis is critical for identifying and mitigating cascading errors that may compromise overall performance. Accordingly, these considerations—ranging from fine-tuning methodologies to error propagation and loss analysis—will be systematically incorporated into the fine-tuning process.

\subsection{Supervised Fine-tuning}
Fine-tuning occurs on two distinct levels: the first involves optimizing a medical domain-adapted language model to improve diagnostic and care delivery capabilities within the core model. The second level focuses on enhancing machine translation through the integration of contextual data, thereby facilitating contextual localization. This multi-level approach not only elevates the model's medical applications but also enables the incorporation of additional languages into the machine translation framework, significantly easing the expansion into other low-resource languages.

\subsubsection{Fine-tuning Medical Domain Adapted Model}
I chose open-source Meditron 70B for its strong performance on medical benchmarks. An A100-80GBx8 GPUs cluster was used to fine-tune the model with a filtered and curated dataset of 520 million medically relevant tokens. Optimization is done using AdamW optimizer and the following hyperparameters:
\[\beta = 0.9,  \epsilon = 1 \times 10^{-5}, \text{ weight decay } = 0.1, \text{ batch size = } 64, \text{ learning rate = } 2 \times 10^{-5}\]
The model was fine-tuned for 2 epochs over 54 hours. The total vocabulary size of the filtered dataset used is 14,227.  

\textbf{Distributed Environment:} I used the Fully Sharded Data Parallel (FSDP) distributed learning framework to distribute the model's parameters across multiple GPUs, thereby reducing the memory footprint per device. 

\textbf{Quantization}: 
Finetuning Meditron 70B on an NVIDIA A100 8-GPU cluster specifically employs Adaptive Weight Quantization (AWQ) to address the dual challenges of managing a 70 billion parameter model's substantial memory requirements and maximizing computational efficiency. By adaptively quantizing the model weights, AWQ significantly reduces the memory footprint and enhances computational speed.

\subsubsection{Fine-tuning Machine Translation Model}
The Seamless M4T v2, an advanced open-source machine translation model from Meta, is pre-trained on a diverse multilingual dataset that includes 101 languages, with a significant portion—over 40\%—categorized as low-resource languages. This model has undergone fine-tuning on an NVIDIA Tesla T4 x 2 GPU using a specialized parallel corpus. This corpus is meticulously curated to include contextually and culturally nuanced medical terms, enabling bidirectional translations between English and four other languages: Telugu, Hindi, Arabic, and Swahili. The fine-tuning process emphasizes the adaptation of the model to accurately reflect the subtleties inherent in culturally sensitive healthcare communication.

\subsubsection{Strategies for Error Propagation Mitigation}
To minimize error amplification in an integrated system that combines a translation model with L2M3
for handling user inputs and generating responses in local languages, I deployed some of the basic techniques:
\begin{itemize}
    \item \textbf{Cross-Lingual Embeddings:} Implemented cross-lingual embeddings to uphold semantic consistency across English, Telugu, Hindi, Swahili, and Arabic, ensuring L2M3's query processing is rooted in precise interpretation.
    \item \textbf{Context Retention Enhancements:} Utilize context retention mechanisms to ensure that
important contextual information is preserved throughout the translation process, thereby
maintaining the integrity of the original query when processed by L2M3.
\end{itemize}

Despite implementing basic techniques to mitigate errors, finding a scalable solution for evaluating
and further reducing error amplification in the system remains challenging.

\section{Evaluation}

Evaluation is done across two distinct but interconnected dimensions, each tailored to affirm the system's utility for CHWs:

\begin{itemize}
    \item \textbf{Domain-Adapted Model L2M3 Performance:} This aspect evaluates the model's capability in accurately interpreting and processing specialized medical terminology, focusing on its precision and comprehension. It assesses the model's effectiveness in diagnostic contexts and its ability to facilitate accurate healthcare delivery, ensuring that its outputs are both clinically relevant and appropriate for real-world medical settings.
    \item \textbf{Translation Model Accuracy:} Precision in translation is scrutinized, evaluating how well the model maintains the integrity of medical information across languages, crucial for clear and error-free communication in diverse linguistic landscapes.

    \end{itemize}

\subsection{Domain-Adapted Model L2M3 Performance}

Traditional evaluation methodologies for domain-adapted models in healthcare have predominantly utilized benchmarks based on medical question-answering (QA) datasets such as PubMedQA and MedQA. While these benchmarks serve as valuable tools for comparative analysis, they may not adequately address the specific needs and nuances of responses associated with healthcare delivery in rural settings.

In response to this need, I propose a novel evaluation framework designed to more effectively assess the applicability and utility of AI-generated responses in the context of rural healthcare. This framework is characterized by its focus on two bespoke evaluation datasets: (1) a set of medical queries relevant to CHWs in rural areas, and (2) a collection of medical guidelines that reflect the practical and situational needs encountered in these settings. These datasets are crafted to encompass both diagnostic inquiries and guideline-related queries, thereby covering a broad spectrum of information needs.

The evaluation process involves the submission of queries from these datasets to the fine-tuned medical model, to generate responses that are then assessed for their accuracy and utility. To ensure a robust and comprehensive evaluation, the generated responses are subjected to validation through two advanced language models: GPT-4 \cite{gpt4systemcard} and Claude Opus \cite{claude3modelcard} APIs. A response is deemed appropriate only if it secures validation (both semantic and contextual similarities) from both models, indicating a high level of reliability and relevance. Conversely, if either API fails to validate the generated response, it is classified as unacceptable.

\begin{figure}[htp]
    \centering
    \includegraphics[scale=0.18]{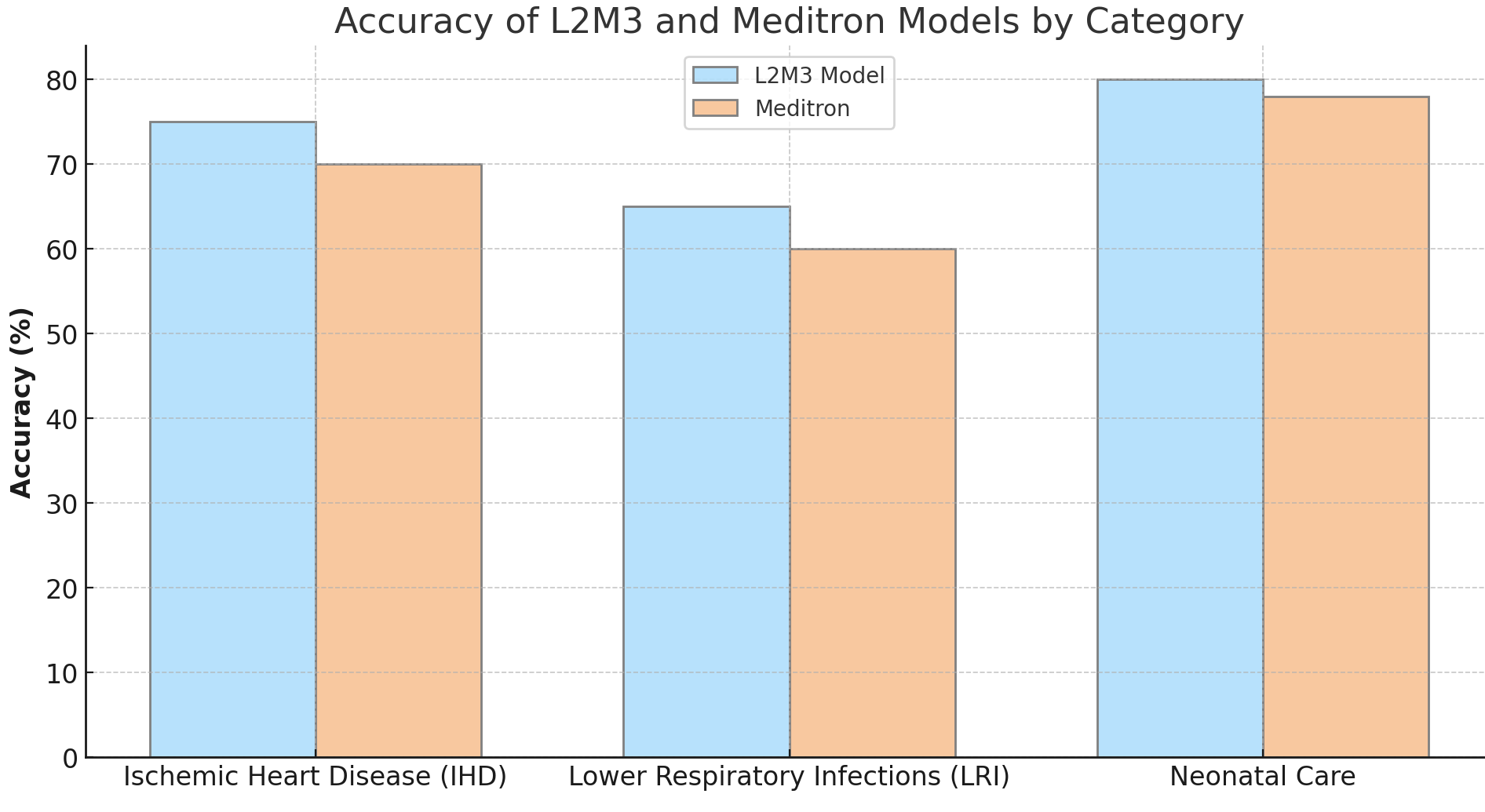}
    \caption{Accuracy of L2M3 and Meditron Models by DALY Top Three Catagories}
    \label{fig:graphic1}
\end{figure}

Figure 3 illustrates the comparative performance of the L2M3 model over Biomistral in terms of user-friendly response generation and the use of easy-to-understand terminology. The bar charts demonstrate a clear advantage of L2M3 in making complex medical information more accessible for CHWs. The enhanced readability and approachability of responses are pivotal for effective communication in healthcare settings.
\subsection{Evaluation of Machine Translation Fine-tuned Model}

The fine-tuning of the machine translation model (Seamless M4T v2 Large) was rigorously evaluated for its translation accuracy, with a particular focus on its proficiency in handling medical terminology and colloquial expressions. The Bilingual Evaluation Understudy (BLEU) metric was used to assess the machine translation efficiency. 

\begin{table}[h!]
\centering
\begin{tabular}{>{\bfseries}lcc}
\toprule
& \multicolumn{2}{c}{\textbf{Seamless M4T v2 Large}} \\
\cmidrule{2-3}
& \textbf{Before Fine-tuning} & \textbf{After Fine-tuning} \\
\midrule
Telugu \textrightarrow{} English & 75.6 & \cellcolor{maxvalue}82.4 \\
Hindi \textrightarrow{} English & 73.4 & \cellcolor{maxvalue}83.1\\
Swahili \textrightarrow{} English & 45.8 & \cellcolor{maxvalue}48.1 \\
Arabic \textrightarrow{} English & 68.5 & \cellcolor{maxvalue}80.5 \\
\midrule
English \textrightarrow{} Telugu & 59.1 & \cellcolor{maxvalue}81.7\\
English \textrightarrow{} Hindi & 62.1 & \cellcolor{maxvalue}83.3 \\
English \textrightarrow{} Swahili & 32.4 & \cellcolor{maxvalue}40.1 \\
English \textrightarrow{} Arabic & 54.3 & \cellcolor{maxvalue}78.9 \\
\bottomrule
\end{tabular}
\vspace{10pt}
\caption{Performance of Seamless M4T Before and After Fine-tuning}
\label{tab:seamless_m4t_finetuning}
\end{table}

As depicted in Table 4, the performance of the fine-tuned translation model demonstrated significant improvement for Indic languages (Hindi, Telugu) and Arabic; however, its enhancement in handling Swahili was comparatively minimal. In the evaluation of translation accuracy, I observed that the translation process is significantly influenced by contextual factors such as gender and age, which occasionally led to errors such as a "2-year-old girl" being inaccurately translated as a "woman." Additionally, the translation of certain medical terms, like "COPD" (Chronic Obstructive Pulmonary Disease), into languages such as Telugu or Hindi, presented notable challenges, indicating difficulties in accurately conveying specialized medical vocabulary across different linguistic contexts. These findings highlight the critical need for contextual awareness and precision in the translation of medical and demographic information.

\section{Integrated L2M3 System }

The system seamlessly integrates fine-tuned components and models to create dependable medical assistants for CHWs. As depicted in Figure 4, CHWs submit information in their local language, which is then translated into English by a precision-tuned translator. This translation undergoes a safety and security check by guardrails. Once cleared, the text is processed by the L2M3 model to generate a response. This response is again vetted by guardrails for quality and safety, then translated back into the CHW's local language before being delivered, ensuring accurate, reliable medical assistance. 

\begin{figure}[htp]
    \centering
    \includegraphics[scale=0.25]{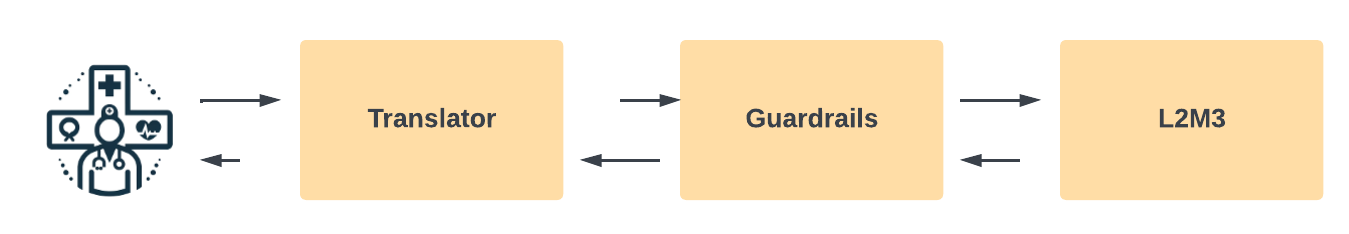}
    \caption{Integrated System for CHW }
    \label{fig:graphic1}
\end{figure}
To guarantee the safety and security of prompts and responses, I implemented NeMo Guardrails, an open-sourced library from NVidia \cite{rebedea-etal-2023-nemo}. Although NeMo's integration with L2M3 and the Translation Model has yet to be assessed, it has been deployed to manage topical relevance, prevent jailbreaking, reduce hallucinations, and ensure secure execution. NeMo Guardrails are designed to maintain the integrity of user queries and the guardrails do not modify user inputs. If a query is detected as ill-formed or unclear, it is returned to the user for clarification, ensuring accuracy in the medical assistance process.

\subsection{Performance of Integrated System}

The integrated model may exhibit reduced performance relative to its individual components, attributed to error propagation arising from the sequential integration of the models.

\begin{table}[h!]
\centering
\begin{tabular}{>{\bfseries}lcc}
\toprule
& \multicolumn{2}{c}{\textbf{Integrated L2M3 Systems }} \\
\cmidrule{2-3}
& \textbf{Before Fine-tuning} & \textbf{After Fine-tuning} \\
\midrule
Telugu \textrightarrow{} English \textrightarrow{} Telugu & 48.1 & \cellcolor{maxvalue}66.4 \\
Hindi \textrightarrow{} English \textrightarrow{} Hindi& 48.1 & \cellcolor{maxvalue}67.3\\
Swahili \textrightarrow{} English \textrightarrow{} Swahili& 27.7 & \cellcolor{maxvalue}42.1 \\
Arabic \textrightarrow{} English \textrightarrow{} Arabic& 43.9 & \cellcolor{maxvalue}64.3 \\

\bottomrule
\end{tabular}
\vspace{10pt}
\caption{Approximate Performance Improvement of Integrated L2M3 System (MT + Fine-tuned Meditron)}
\label{tab:seamless_m4t_finetuning}
\end{table}

Table 5 illustrates the estimated performance enhancement in the integrated system following fine-tuning. It's important to note that these figures are approximations, as potential error propagation from the sequential integration of models may impact actual performance.

\subsection{Truthfulness}
Ensuring accuracy and mitigating misinformation are paramount in language model applications to prevent hallucinations. For this evaluation, the MedHALT dataset \cite{umapathi2023medhalt}, featuring Reasoning Hallucination Tests (RHT) and Memory Hallucination Tests (MHT), was employed. Focusing on reasoning-based tests due to their relevance and applicability, the datasets were translated into Hindi, Telugu, Arabic, and Swahili using Azure Cognitive Translation APIs. 
some of the translated datasets were refined and contextual enriched to ensure their relevance and applicability across the targeted languages.

The tests are designed to evaluate a language model's capability to perform logical reasoning with medical data and to produce outputs that are not only coherent but also factually correct. Specifically, they ensure the model does not fabricate information. The tests encompass three distinct types:
\begin{itemize}
    \item \textbf{False Confidence Testing (FCT):} This evaluates the model's inclination to provide answers with undue confidence, particularly when it does not have adequate information.
    \item \textbf{None of the Above (NOTA) Test:} This determines the model's proficiency in recognizing and disregarding irrelevant or incorrect information.

    \item 
\textbf{Fake Questions Test (FQT):} This involves challenging the model with spurious or absurd medical queries to assess its capability to accurately detect and process such inputs.
\end{itemize}

\textbf{Metrics Used \cite{umapathi2023medhalt}:}

\begin{equation}
\text{Accuracy} = \frac{\text{Number of Correct Predictions}}{\text{Total Number of Predictions}}
\end{equation}

Given a set of predictions, the final score $S$ can be calculated using the formula:
\begin{equation}
S = \frac{1}{N} \sum_{i=1}^{N} \left( I(y_i = \hat{y}_i) \cdot P_c + I(y_i \neq \hat{y}_i) \cdot P_w \right)
\end{equation}

Where:
\begin{itemize}
    \item $S$ is the final score,
    \item $N$ is the total number of samples,
    \item $y_i$ is the true label of the $i$-th sample,
    \item $\hat{y}_i$ is the predicted label of the $i$-th sample,
    \item $I(\text{condition})$ is the indicator function that returns 1 if the condition is true and 0 otherwise,
    \item $P_c$ is the points awarded for a correct prediction, and
    \item $P_w$ is the points deducted for an incorrect prediction.
\end{itemize}

\begin{table}[htbp]

\label{tab:performance}
\centering
\begin{tabularx}{\textwidth}{@{}Xcccccc@{}}
\toprule
\textbf{Model} & \multicolumn{2}{c}{\textbf{ FCT}} & \multicolumn{2}{c}{\textbf{FQT}} & \multicolumn{2}{c}{\textbf{NOTA}} \\
\cmidrule(l){2-7} 
      & \textbf{Accuracy} & \textbf{Score} & \textbf{Accuracy}  & \textbf{Score} & \textbf{Accuracy} & \textbf{Score} \\
\midrule
GPT-3.5*         & 34.15 & 33.37 & 71.64 & 11.99 & 27.64 & 18.01 \\
Llama-2 70B*     & \cellcolor{maxvalue}42.21 & \cellcolor{maxvalue}52.38 & \cellcolor{maxvalue}97.26 &\cellcolor{maxvalue} 17.94 & 77.53 & 188.66 \\
\midrule
Integrated L2M3 - Telugu    & 36.3 & 34.35 & 87.41 & 16.1 & 79.5 & 191.6 \\
Integrated L2M3 - Hindi    & 39.17 & 43.12 & 89.66 & 18.2 & \cellcolor{maxvalue}82.34 &\cellcolor{maxvalue} 193.85 \\
Integrated L2M3 - Arabic    & 34.25 & 33.32 & 88.41 & 17.32 & 79.41 & 190.79 \\
\bottomrule
\end{tabularx}
\vspace{6pt}
\caption{Evaluation results of LLMs and Integrated System on Reasoning Hallucination Tests. \\ * Evaluation was conducted as part of the work \cite{umapathi2023medhalt} 
P\_c = 1, \quad P\_w = -0.25}
\end{table}

The Integrated L2M3 system outperformed GPT-3.5 in Reasoning Hallucination Tests (RHT), demonstrating superior effectiveness despite encountering challenges such as translation inaccuracies and error amplification. Additionally, discrepancies within the datasets, particularly in the Reasoning False Confidence Testing (FCT), affected some evaluation tests across all three languages. Furthermore, hallucination evaluations for Swahili were not feasible due to validation check failures in the translated data.

\section{Conclusion}
In conclusion, this paper introduces an innovative strategy to combat the projected global shortfall of health workers, with a focus on LMICs. Through the integration of LLMs with advanced machine translation technologies, I've created a customized Multilingual Language Model that empowers CHWs with limited orientation. This approach transcends linguistic and cultural barriers, significantly improving healthcare service accessibility and quality by equipping CHWs with contextually relevant medical knowledge and diagnostic tools.

The system's modular design is a critical feature, enabling rapid adaptation across diverse linguistic and cultural settings while leveraging open-source components to reduce operational costs substantially. Importantly, the composability of the solution is key, not just for its scalable and effective rollout across multiple countries, but for its unique ability to facilitate independent enhancements in the performance of local languages. This aspect is crucial in ensuring that medical assistance provided by this system is both precise and culturally relevant, further mitigating healthcare disparities.

This work highlights the potential of AI in bridging the healthcare workforce gap in LMICs and enhancing global health outcomes. It presents a scalable and adaptable model for leveraging AI to address healthcare challenges, emphasizing the importance of language-specific performance improvements

\bibliography{references} 
\bibliographystyle{plain}
\end{document}